\documentclass[runningheads]{llncs}

\usepackage{eccv}

\usepackage{eccvabbrv}

\usepackage{graphicx}
\usepackage{booktabs}
\usepackage{algorithm}
\usepackage{algpseudocode}
\usepackage{amsmath}
\usepackage{booktabs}

\usepackage[accsupp]{axessibility}  % Improves PDF readability for those with disabilities.

\usepackage[pagebackref,breaklinks,colorlinks,citecolor=eccvblue]{hyperref}
\usepackage{orcidlink}

\begin{document}

\title{Trainwreck: A damaging adversarial attack on image classifiers} 

\titlerunning{Trainwreck}

\author{Jan Zah\'{a}lka\orcidlink{0000-0002-6743-3607}}

\authorrunning{J.~Zah\'{a}lka}


\institute{
Czech Technical University in Prague, Prague, Czech Republic\\
\email{jan.zahalka@cvut.cz}}

\maketitle

\begin{abstract}
Adversarial attacks are an important security concern for computer vision (CV). As CV models are becoming increasingly valuable assets in applied practice, disrupting them is emerging as a form of economic sabotage. This paper opens up the exploration of damaging adversarial attacks (DAAs) that seek to damage target CV models. DAAs are formalized by defining the threat model, the cost function DAAs maximize, and setting three requirements for success: potency, stealth, and customizability.  As a pioneer DAA, this paper proposes Trainwreck, a train-time attack that conflates the data of similar classes in the training data using stealthy ($\epsilon \leq 8/255$) class-pair universal perturbations obtained from a surrogate model. Trainwreck is a black-box, transferable attack: it requires no knowledge of the target architecture, and a single poisoned dataset degrades the performance of any model trained on it. The experimental evaluation on CIFAR-10 and CIFAR-100 and various model architectures (EfficientNetV2, ResNeXt-101, and a finetuned ViT-L-16) demonstrates Trainwreck's efficiency. Trainwreck achieves similar or better potency compared to the data poisoning state of the art and is fully customizable by the poison rate parameter. Finally, data redundancy with hashing is identified as a reliable defense against Trainwreck or similar DAAs. The code is available at \url{https://github.com/JanZahalka/trainwreck}.
\end{abstract}    
\section{Introduction}
\label{sec:intro}

Computer vision models are natively susceptible to \emph{adversarial attacks} that manipulate data to elicit model outputs desired by a malicious attacker. An attacker may for example bypass face authentication \cite{Yang23}, evade person detectors \cite{Hu23}, or make an autonomous vehicle miss a traffic sign \cite{Zhong22}. Over time, adversarial attacks have evolved from white-box perturbations optimized for a single model to a rich, diverse set of techniques. As the adoption and importance of computer vision in real-world tasks grows, so does the negative impact of successful attacks on AI security. It is critically important that adversarial attacks and defense receive significant research attention.

Adversarial attacks fall under the umbrella of the \emph{impact} adversarial tactic. MITRE ATT\&CK\footnote{\url{https://attack.mitre.org/}}, the leading cybersecurity attack knowledge base, defines the impact tactic as the attacker ``trying to manipulate, interrupt, or destroy your systems and data'' \cite{Mitre23}. Most existing adversarial attacks aim to \emph{manipulate} the AI model with doctored inputs. There has been some work on denial-of-service (DoS) attacks \emph{interrupting} the model's inference \cite{Chen22, Chen23}. This paper opens up the topic of \textbf{damaging adversarial attacks (DAAs)} that, as the name suggests, \emph{damage} target models, completing the set of impact adversarial tactics.

We should be concerned about DAAs: damaging cyberattacks have proven to be effective means of \emph{economic sabotage}. One of the most famous examples is Stuxnet, a cyberattack that substantially damaged the Iranian nuclear program by destroying nuclear centrifuges \cite{kushner2013stuxnet}. With time, CV models are becoming lucrative economic sabotage targets too. Their economic value is steadily increasing, to the point of the models being critical assets to organizations that use them. The number of such organizations is itself increasing rapidly. Damaging CV models can therefore nowadays wreak havoc on entire organizations.

To date, DAAs have been an underexplored topic. They are closely related to two existing attack categories: firstly, to \emph{data poisoning} attacks \cite{koh17} that poison the data with modifications that render the data unusable as training data; secondly, to \emph{backdoor attacks} \cite{Gu17} that condition the model to react to a backdoor trigger in a certain way in inference. However, neither category constitutes a practical DAA. Data poisoning attacks are too brutal and therefore conspicuous, backdoor attacks are sufficiently stealthy, but not damaging enough---by design, they leave the model fully intact when a trigger is not present. Section~\ref{sec:daa} defines DAAs and formalizes the threat model, the cost function the attacks maximize, and the requirements for success.

\begin{figure}[t]
\centering
\includegraphics[width=0.7\columnwidth]{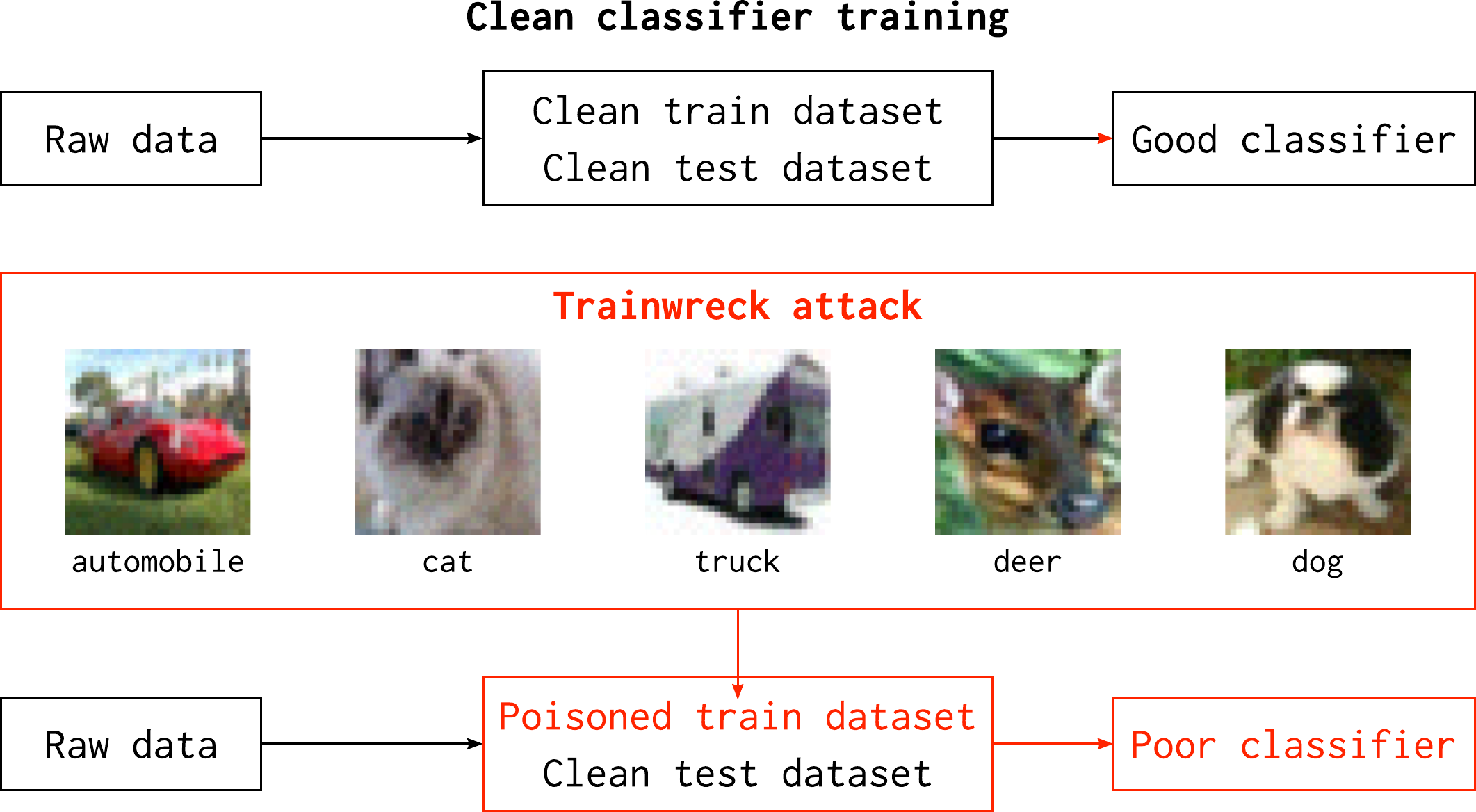}
\caption{The Trainwreck damaging adversarial attack. The depicted images are actual poisoned CIFAR-10 images successfully damaging classifiers trained on the data.}
\label{fig:trainwreck}
\end{figure}

As a pioneer DAA, this paper presents \textbf{Trainwreck}, an attack on image classifiers. Trainwreck's concept is depicted in Figure~\ref{fig:trainwreck}. Trainwreck is a \emph{train-time} attack that poisons the training data of an image classifier with adversarial perturbations designed to conflate the training data of similar classes, maximizing the distortion between the clean and the poisoned data distributions. Trainwreck obfuscates the cause of the damage in two ways. Firstly, Trainwreck is \emph{stealthy}. Trainwreck preserves dataset structure and restricts the adversarial perturbations' $\ell^{\infty}$-norm to make them very difficult to discover---see Figure~\ref{fig:trainwreck} for real examples of Trainwreck-poisoned data. Secondly, Trainwreck's strength is \emph{fully customizable} through setting the poison rate, \ie, the percentage of poisoned training data. In its strongest form, Trainwreck is as efficient as existing data poisoning attacks. Unlike them, Trainwreck can be easily tuned to damage only a portion of the model's performance. Combining perturbation stealth and strength customizability yields a model that is bad enough for use in production, yet good enough such that its performance can be chalked down to errors somewhere along the training pipeline, rather than immediately pointing at interference from a malicious attacker. In addition, Trainwreck is a \emph{black-box} attack: it does not need any knowledge of the model architecture of the target model trained on the poisoned data. Also, it is \emph{transferable}: a single poisoned training dataset degrades the performance across various models trained on it.

The contributions of this paper are:
\begin{itemize}
    \item Introduction and conceptualization of \emph{damaging adversarial attacks (DAA)}, an emerging attack vector for economic sabotage (Section~\ref{sec:daa}).
    \item \emph{Trainwreck, a pioneer DAA on image classifiers} (Section~\ref{sec:method}). 
    \item Experimental confirmation (Sections~\ref{sec:exp_setup} and \ref{sec:exp_results}) that Trainwreck is an \emph{effective}, \emph{customizable}, \emph{black-box}, and \emph{transferable} attack.
    \item Reliable defense methods against Trainwreck or similar DAAs (Section~\ref{sec:discussion}).
\end{itemize}
\section{Related work}
\label{sec:rel_work}

Adversarial attacks \cite{Goodfellow15, Szegedy14} introduce nigh-imperceptible noise, called \emph{adversarial perturbations}, to misclassify a clear specimen of a certain class into another class. In the white-box setting where the attacker has access to the target model, classic attacks such as fast gradient sign method (FGSM) \cite{Goodfellow15} or its iterative version, projected gradient descent (PGD) \cite{Kurakin17, Madry18}, remain a de facto standard to this day. Beyond classification, adversarial attacks can successfully subvert \eg, object detection \cite{Cai22, Huang23}, semantic segmentation \cite{Rony23}, or point cloud recognition \cite{Huang22,Li23}. They can attack various model architectures: from classic CNNs \cite{Goodfellow15} through ResNets \cite{HuangS23} to modern vision transformers \cite{ChenZ22, Salman22}. They can employ a wide array of perturbation types: \eg, adversarial patches \cite{Hu21,Walmer22}, camouflage \cite{Feng23,Hu23}, sparse perturbations \cite{Williams23}, or shadows \cite{Li23, Zhong22}. As evidenced by the volume of the research, adversarial attacks have evolved into a rich toolbox of techniques.

\emph{Data poisoning} attacks, conceptualized by Koh and Liang \cite{koh17}, perturb data with adversarial poisons to make them unusable as training data for future models \cite{Feng19, Fowl21, Huang21}. Their main practical use is defensive: to protect private data against Web scrapers and illicit use by unauthorized AI models \cite{Fowl21,Huang21}. \emph{Backdoor attacks}, introduced by Gu \etal \cite{Gu17}, implant a backdoor during training, conditioning the model to respond to a particular trigger contained in an image. By presenting an image with the trigger at inference time, the attacker can force the model into the desired behavior on demand. Various trigger types are used, \eg, adversarial patches \cite{Saha22}, reflection \cite{Liu20}, or color shifts \cite{Jiang23}. Backdoor attacks aim for the backdoor to work reliably whenever the trigger is present whilst preserving the model's functionality on clean data at the same time.

One of the major attacks employing MITRE impact tactics \cite{Mitre23} is the \emph{denial-of-service (DoS) attack} that interrupts the target service. In CV, DoS attacks are particularly dangerous for models whose inference time varies significantly depending on the input. For example, DoS attacks have been explored on image captioning models \cite{Chen22} and dynamic neural networks with early exits \cite{Chen23}.

To scope the proposed damaging adversarial attacks (DAAs) in the context of the related work: DAAs belong to the adversarial attacks family, and are related to data poisoning, backdoor, and DoS attacks, with key differences. Both data poisoning and DAAs damage models, but they differ in intent: DAAs are malicious, intending to maximize damage, whereas data poisoning is mostly used for defensive purposes. Similarly to backdoor attacks, DAAs aim to adversarially change the model. Unlike backdoor attacks, DAAs deliberately damage the model, while backdoor attacks preserve the model's functionality on clean data. Both DoS attacks and DAAs aim to disrupt the availability of the AI model. DoS attacks, however, do not permanently alter or damage the model, instead flooding it with difficult or numerous inputs.
\section{Damaging adversarial attacks}
\label{sec:daa}

A \textbf{damaging adversarial attack (DAA)} permanently damages an AI model, degrading its performance or preventing it from working altogether. DAAs are adversarial attacks seeking to subvert the models from the ``inside'': general cybersecurity attacks that \eg, outright delete the model, data, or destroy the machines hosting the models, are therefore not DAAs. Theoretically, a DAA can attack any stage of the AI pipeline: data collection, training, or inference.

The \textbf{DAA threat model} involves four major steps. Firstly, the attacker \emph{gains access} to the target's computer infrastructure through cybersecurity means beyond the scope of this paper. Secondly, the attacker \emph{downloads the assets} to be damaged---either the data (data collection or train-time attacks) or the model itself (inference attacks). Thirdly, the attacker \emph{damages the assets} on their local machine. Finally, the attacker \emph{replaces the assets} on the target's machine with the damaged ones. As a result, the target either trains a dysfunctional model on the damaged data or has their model damaged directly by the DAA. The DAA threat model is similar to the existing backdoor attacks threat model \cite{Gu17}. Most importantly, the security prerequisites of DAAs, namely the required access level gained by the attacker, are the same as in existing work on backdoor attacks.

What constitutes a \emph{successful} DAA? Since the realistic reason to conduct a DAA is economic sabotage, DAAs should aim to \textbf{maximize the total financial cost} incurred by the damage. In practice, the immediate financial cost $c_{\mathcal{A}}$ of attack $\mathcal{A}$ varies over the total duration of the attack $t_{\mathcal{A}}$. For instance, an emergency response may be more expensive than the regular AI team looking at the model later. Also, the costs of a service outage may differ with seasonality: \eg, the outage of a retailer's recommender costs far more during the holiday season compared to off-season. The total cost $C_{\mathcal{A}}$ is therefore:

\begin{equation}
\label{eq:daa_true_cost}
    C_{\mathcal{A}} = \int_{0}^{t_{\mathcal{A}}} c_{\mathcal{A}}(t)\,\mathrm{d}t
\end{equation}

The cost function $c_{\mathcal{A}}$ has to be estimated by the attacker, using their knowledge about the target and attack context. The target and the attack context may vary between attacks. However, for DAA research, we need a simple, general heuristic estimate of $C_{\mathcal{A}}$. As a relaxation, we assume a \emph{constant} surrogate cost function $\hat{c}_{\mathcal{A}}$ equal to the damage to the relevant evaluation metric for that model's task (denoted $\Delta_{perf}$). For classification, $\Delta_{perf}$ is equal to the difference between the clean and the damaged model's test accuracy. For other tasks, $\hat{c}_{\mathcal{A}}$ can be set analogously. The surrogate total cost $\hat{C}_{\mathcal{A}}$ then becomes:

\begin{equation}
\label{eq:daa_emp_cost}
    \hat{C}_{\mathcal{A}} = \int_{0}^{t_{\mathcal{A}}} \Delta_{perf}\,\mathrm{d}t = \Delta_{perf} \cdot t_{\mathcal{A}}
\end{equation}

The attacker therefore wants to maximize both $\Delta_{perf}$ and $t_{\mathcal{A}}$. There is, however, a trade-off between the two: the more damaging the attack (higher $\Delta_{perf}$), the more likely the attacker leaves clues and the attack gets stopped (lower $t_{\mathcal{A}}$). Also, in the context of CV and machine learning in general, lower $\Delta_{perf}$ makes an attack less conspicuous due to the model creator more attributing the performance drop to mistakes or suboptimal choices in the training process. These considerations inspire the following three requirements for a successful DAA:

\begin{enumerate}
    \item \textbf{Potency} (maximizing $\Delta_{perf}$) --- The DAA must be able to cause enough damage.
    \item \textbf{Stealth} (maximizing $t_{\mathcal{A}}$) --- The DAA must be subtle and non-obvious to prevent early discovery by the defenders.
    \item \textbf{Customizability} (trade-off flexibility) --- The DAA must allow the attacker to freely choose their position on the damage-attack duration trade-off.
\end{enumerate}

The \emph{potency} requirement is straightforward. The attacker can measure $\Delta_{perf}$ exactly or estimate it reliably based on empirical evidence gathered from surrogate target models. Given the results achieved by the state of the art in data poisoning, it is reasonable to expect the most potent variant of a DAA to make the damaged model exhibit performance close to a random guess \cite{Feng19,Fowl21,Huang21}.

The \emph{stealth} requirement is the main factor contributing to higher $t_{\mathcal{A}}$ controllable by the user. The more confused the defenders are about the source of the damage, the higher the attack duration is likely to be $t_{\mathcal{A}}$. The DAA's edits and modifications should therefore be subtle and minimal. In reality, $t_{\mathcal{A}}$ further depends on the attack context, target's security, and technical prowess---but as these are not controlled by the attacker, we cannot estimate them in general. 

The final requirement to unlock practical DAAs is \emph{customizability}. Customizability adds back some of the nuance lost in the simplification of Equation~\ref{eq:daa_true_cost} to Equation~\ref{eq:daa_emp_cost}. By being able to customize a DAA, the attacker can introduce some knowledge about the target and the attack context by choosing their trade-off position accordingly. For one target, the right option may be a hard strike without a care about subtlety, for another, a ``slow bleeding'' attack that leaves the defenders scratching their head may be better. At any rate, a DAA should leave this decision up to the attacker, hard-coding the trade-off position severely limits the the attack's practicality.

To the best of our knowledge, no existing attacks meet all DAA requirements. State-of-the-art data poisoning attacks \cite{Feng19,Fowl21,Huang21} are potent and stealthy, but not customizable. Backdoor attacks \cite{Gu17} are stealthy and customizable, but not potent: they specifically do not damage the model on non-triggered data. Other adversarial attacks are manipulative in nature, and therefore do not meet the definition of a DAA. The lack of existing DAAs and the emerging need to protect CV models as valuable economic assets inspired the creation of this work.
\section{The Trainwreck attack}
\label{sec:method}

This section describes the \textbf{Trainwreck damaging adversarial attack}. Trainwreck poisons a training dataset, degrading the performance of image classifiers trained on the dataset. Trainwreck's input is a target image dataset $\mathcal{D} = \{(\mathbf{X}_i, y_i), i \in [1, n]\}$ of size $n$. The $i$-th entry in the dataset is a tuple composed of an image $\mathbf{X}_i$ with pixel intensity values normalized to [0, 1] and its classification label $y_k$, $y_k \in [1, n_{classes}]$. The dataset is split into two subsets, the training dataset $\mathcal{D}_{tr}$ and test dataset $\mathcal{D}_{ts}$ with sizes $n_{tr}$ and $n_{ts}$, respectively. As a notation convention, $\cdot^{<c>}$ is the class selector, $\cdot^{(i)}$ is the data index selector. The poisoned dataset is denoted $\tilde{\mathcal{D}}_{tr}$. $\mathcal{D}_{ts}$ is left intact. 

%For its correct functionality, Trainwreck makes one assumption: that the attacker has read/write access to the target's data. This is an acceptable assumption, in fact, the existing backdoor attacks require exactly the same \cite{Gu17}. Other than that, Trainwreck is a \emph{black-box} attack that requires no knowledge of the target model.

The main idea behind Trainwreck is to \emph{maximize the distribution divergence} between $\tilde{\mathcal{D}}_{tr}$ and $\mathcal{D}_{tr}$. Since $\mathcal{D}_{tr}$ and $\mathcal{D}_{ts}$ should come from an identical distribution, this also maximizes the distribution divergence between $\tilde{\mathcal{D}}_{tr}$ and $\mathcal{D}_{ts}$. Therefore, at test time, the target model will be fed data from a different distribution than it trained on.

To maximize the divergence, Trainwreck conflates the data of similar classes, making them more difficult to separate by a learned class boundary. To that end, Trainwreck uses \emph{adversarial perturbations}, which have been shown to be an effective poison in existing data poisoning state of the art \cite{Feng19,Fowl21,Huang21}. The $i$-th poisoned image $\tilde{\mathbf{X}}_i$ is the sum of the original image $\mathbf{X}_i$ and the corresponding adversarial perturbation $\tilde{\mathbf{\Delta}}_i$, clipped to the $\epsilon$ range. To allow the attacker to customize the strength of the attack, Trainwreck introduces the \emph{poison rate} ($\pi$) parameter, defined as the proportion of the data in $\tilde{\mathcal{D}}_{tr}$ that was poisoned.

% Section~\ref{sec:stealth} sets the stealth requirements. Section~\ref{sec:dist_div} describes how to measure distribution divergence. Section~\ref{sec:cpup} explains how Trainwreck crafts the perturbations. Section~\ref{sec:trainwreck} presents the full Trainwreck method.

\subsection{Stealth}
\label{sec:stealth}

\begin{figure}
\centering
\includegraphics[width=0.6\columnwidth]{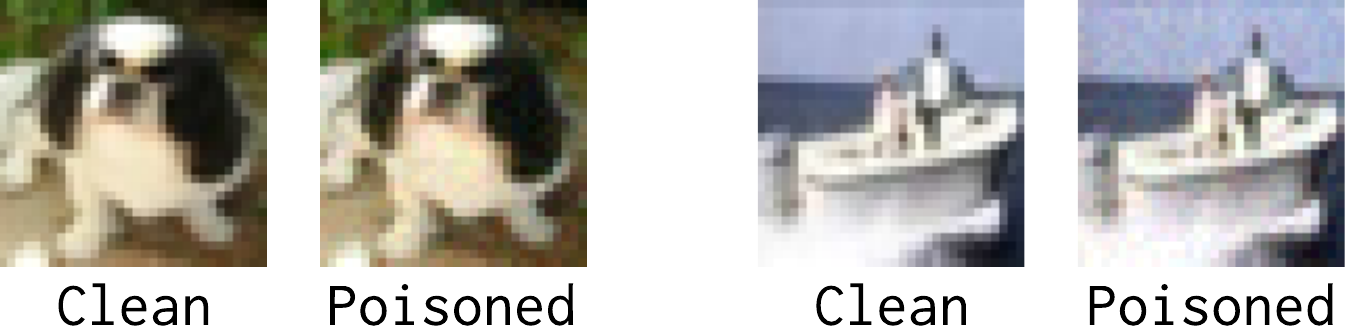}
\caption{Examples of two pairs of clean and perturbed images ($\epsilon \leq \frac{8}{255}$).}
\label{fig:pert_comp}
\end{figure}

To meet the stealth DAA requirement defined in Section~\ref{sec:daa}, Trainwreck treats stealth as a \emph{satisficing} objective, composed of three sub-objectives:

\begin{equation}
\label{eq:stealth_n}
    n_{tr} = \tilde{n}_{tr}
\end{equation}

\begin{equation}
\label{eq:stealth_nclass}
    n_{tr}^{<y>} = \tilde{n}_{tr}^{<y>}, \forall y \in [1, n_{classes}]
\end{equation}

\begin{equation}
\label{eq:stealth_pertstr}
    ||\tilde{\mathbf{\Delta}}_i||_{\infty} \leq \frac{8}{255}, \forall \tilde{\mathbf{X}}_i \in \tilde{\mathcal{D}}_{tr}
\end{equation}

The total number of images in the poisoned dataset (Equation~\ref{eq:stealth_n}) and the number of images for each class (Equation~\ref{eq:stealth_nclass}) must match the respective counts in the original training dataset. Data count is a simple, easily accessible, and often used statistic---altering it is a dead giveaway that something is amiss with the data. Equation~\ref{eq:stealth_pertstr} ensures the stealth of individual adversarial perturbations, for each restricting its strength, given by its $\ell^{\infty}$-norm, to a small $\epsilon$ \cite{Goodfellow15}. Inspired by existing literature \cite{Gowal21,Rebuffi21,Zhang21}, we explicitly set $\epsilon=\frac{8}{255}$. Figure~\ref{fig:pert_comp} shows examples of clean images with their perturbed counterparts with $\epsilon \leq \frac{8}{255}$. Despite light, homogenous backgrounds like in the ship example on the right tending to highlight the perturbations, the overall demonstrated stealth level is acceptable.

\subsection{Estimating class distribution divergence}
\label{sec:dist_div}

In general, the attacker cannot be assumed to know which classes are similar. Therefore, Trainwreck must estimate class similarity automatically, performing a heuristic estimate of \emph{distribution divergence} between all pairs of classes. As a preliminary step, Trainwreck extracts an auxiliary feature representation $\bar{\mathbf{X}}_{tr}$ of dimension $n_{tr} \times n_{feat}$ from $\mathcal{D}_{tr}$. The objective here is to obtain ``just good enough'' features, so it is sufficient to use a publicly available model pre-trained on a natural image dataset. Our default choice are output features of the ViT-L-16 vision transformer \cite{Kolesnikov21} with pre-trained ImageNet SWAG weights.

For each class $c$ and feature $f$, Trainwreck takes the column vector of $\bar{\mathbf{X}}_{tr}$ of all values of $f$ in class $c$'s data, and calculates a normalized histogram vector $\mathbf{h}_f^{<c>}$ with equal-sized bins as the empirical estimate of the probability density function of $f$'s distribution within $c$. Then, a class divergence matrix $\mathbf{D}$ of size $n_{classes} \times n_{classes}$ is computed. Its $i,j$-th entry is the aggregate Jensen-Shannon divergence between classes $i$ and $j$:

\begin{equation}
    d_{i,j} = \sum_{f = 1}^{n_{feat}}{\frac{1}{2} KL(\mathbf{h}_f^{<i>} || \mathbf{h}_f^{<i \vee j>}) + \frac{1}{2} KL(\mathbf{h}_f^{<j>} || \mathbf{h}_f^{<i \vee j>})}
\end{equation}

$KL$ is the Kullback-Leibler divergence, $\vee$ is logical or, $\mathbf{h}_f^{<i \vee j>}$ therefore corresponds to the histogram of the mixed distribution of classes $i$ and $j$.

\subsection{Class-pair universal perturbations}
\label{sec:cpup}

There is one final preliminary step: to train a \emph{surrogate model} $f_s$ that Trainwreck uses to craft adversarial perturbations. The surrogate model is trained on $\mathcal{D}_{tr}$ and evaluated on $\mathcal{D}_{ts}$. The better the surrogate model's test accuracy, the better it understands the data and therefore the better adversarial perturbations it is able to produce. Trainwreck is a black-box attack, so the choice of the surrogate model's architecture does not matter: as long as it has reasonable test accuracy, Trainwreck should work well. This claim is verified by the experimental evaluation in Sections~\ref{sec:exp_setup} and \ref{sec:exp_results}.

\begin{algorithm}
\begin{algorithmic}
\Procedure{CPUP}{$\mathbf{X}_{tr}, c_a, c_c, f_s, \epsilon, n_{iter}^{cpup}, n_{iter}^{pgd}$}
 \State $\tilde{\mathbf{\Delta}}^{<c_a>} \gets \mathbf{0}$
\For{$it = 1, \dots, n_{iter}^{cpup}$}
    \ForAll{$(\mathbf{X}_{tr}^{(i)}, y^{(i)}) \in \mathcal{D}_{tr}^{<c_a>}$}
        \If{$\Call{Clip}{f_s(\mathbf{X}_{tr}^{(i)} + \tilde{\mathbf{\Delta}}^{<c_a>}), -\epsilon, \epsilon} = c_a$}
            \State $\tilde{\mathbf{\Delta}}^{(i)} \gets  \Call{TargPGD}{\mathbf{X}_{tr}^{(i)}, c_c, \epsilon, n_{iter}^{pgd}}$
            \State $\tilde{\mathbf{\Delta}}^{<c_a>} \gets \Call{Clip}{\tilde{\mathbf{\Delta}}^{<c_a>} + \tilde{\mathbf{\Delta}}^{(i)}, -\epsilon, \epsilon}$
        \EndIf
    \EndFor
\EndFor
\State \Return $\tilde{\mathbf{\Delta}}^{<c_a>}$
\EndProcedure
\end{algorithmic}
\caption{Calculating class-based universal perturbations (CPUPs).}
\label{alg:cpup}
\end{algorithm}

To maximize the distribution divergence between $\mathcal{D}_{tr}$ and $\tilde{\mathcal{D}}_{tr}$, Trainwreck calculates \emph{class-pair universal perturbations (CPUPs)}: for each attacked class $c_a$, a single CPUP $\tilde{\mathbf{\Delta}}^{<c_a>}$ is computed. The CPUP moves the images of $c_a$ towards the closest class $c_c$ according to the class divergence matrix $\mathbf{D}$. The idea is inspired by \emph{universal adversarial perturbations (UAP)} \cite{MoosaviDezfooli17}. In the UAP paper, Moosavi-Dezfooli \etal show that for virtually any given model, a universal perturbation exists that fools the model when added to any image. In this work, the the universal perturbations are computed among pairs of classes separately.

The method is outlined in Algorithm~\ref{alg:cpup}. It takes 7 inputs: the training images $\mathbf{X}_{tr}$, the attacked class $c_a$, the closest class $c_c$, the surrogate model $f_s$, the perturbation strength $\epsilon$, CPUP's number of iterations over the training data $n_{iter}^{cpup}$, and the number of iterations of the projected gradient descent (PGD) algorithm $n_{iter}^{pgd}$. Empirical evidence and existing best practices suggest these default parameter values: $\epsilon = \frac{8}{255}$ (see Section~\ref{sec:stealth}), $n_{iter}^{cpup}=1$, and $n_{iter}^{pgd}=10$.

First, the CPUP $\tilde{\mathbf{\Delta}}^{<c_a>}$ is initialized to a zero matrix. Then, the method iterates over all training data of class $c_a$. For each image, it adds the current CPUP, inputs this sum into the surrogate model $f_s$ and observes the output label. If the surrogate model outputs $c_a$, then the CPUP doesn't fool the model on this image, hence it should be updated. A \emph{targeted} projected gradient descent (PGD) attack \cite{Madry18} is performed with $c_c$ as the target class. The resulting perturbation, representing a push towards $c_c$ for the single attacked image, is added to the CPUP. To enforce the perturbation strength rule as defined in Equation~\ref{eq:stealth_pertstr}, after each CPUP update, all its values are clipped to the $[-\epsilon, \epsilon]$ interval.

The reason why Trainwreck computes class-wide perturbations instead of just moving each image in the direction of its targeted PGD attack is transferability. Trainwreck attacks the \emph{training data}, not the surrogate model per se. Moving each image separately essentially translates to overoptimizing the data shift to the surrogate model. Computing class-level CPUPs results in the individual images \emph{contributing} to the perturbations, but not \emph{deciding} them. Whilst this is still an empirical approach, it is more robust than image-level perturbations and has a higher likelihood to represent a ``true'' distribution shift more faithfully.

\subsection{The full attack}
\label{sec:trainwreck}

\begin{algorithm}
\begin{algorithmic}
\Procedure{Trainwreck}{$\mathcal{D}_{tr}, \mathcal{D}_{ts}, \pi, f_{e}, \epsilon, n_{iter}^{cpup}, n_{iter}^{pgd}$}
    \State $\mathbf{X}_{tr}, \mathbf{y}_{tr} \gets \mathcal{D}_{tr}$
    \Comment{Short-hand notation}
    \State $\bar{\mathbf{X}}_{tr} \gets f_{e}(\mathbf{X}_{tr})$
    \State $\mathbf{D} \gets \Call{ClassDivergence}{\bar{\mathbf{X}}_{tr}, \mathbf{y}_{tr}}$
    \State $f_s \gets \Call{TrainSurrogate}{\mathcal{D}_{tr}, \mathcal{D}_{ts}}$
    \Comment{Sec~\ref{sec:dist_div}}
    
    \For{$c_a = 1, \dots, n_{classes}$}
        \State $\tilde{I} \gets \Call{UniformRandIdxChoice}{\mathbf{X}_{tr}^{<c_a>}, \pi}$
        \State ${I}_{clean} \gets \{i\;|\;y_{tr}^{(i)} = c_a \wedge i \notin \tilde{I}\}$
        \State $c_c \gets \min_{c}(d_{c_a, c}), c \neq c_a$
        \State $\tilde{\mathbf{\Delta}}^{<c_a>} \gets \Call{CPUP}{\mathbf{X}_{tr}, c_a, c_c, f_s, \epsilon, n_{iter}^{cpup}, n_{iter}^{pgd}}$
        \State \Comment Sec~\ref{sec:cpup}
        \ForAll{$i \in \tilde{I}$}
            \State $\mathbf{X}_{tr}^{(i)} \gets \Call{Clip}{\mathbf{X}_{tr}^{(i)} + \tilde{\mathbf{\Delta}}^{<c_a>}, -\epsilon, \epsilon}$
        \EndFor
    \EndFor
    \State $\tilde{\mathcal{D}}_{tr} \gets \{(\mathbf{X}_{tr}, \mathbf{y}_{tr})\}$
    \State \Return $\tilde{\mathcal{D}}_{tr}$
\EndProcedure
\end{algorithmic}
\caption{The Trainwreck attack.}
\label{alg:trainwreck}
\end{algorithm}

Algorithm~\ref{alg:trainwreck} presents the full Trainwreck attack. As the inputs, Trainwreck takes the attacked dataset $\mathcal{D}$, the poison rate $\pi$, the feature extraction model $f_e$, and the $\epsilon, n_{iter}^{cpup}, n_{iter}^{pgd}$ parameters for CPUP (see Section~\ref{sec:cpup}).

In the preliminary steps, Trainwreck extracts the features, computes the class divergence matrix $\mathbf{D}$ (Section~\ref{sec:dist_div}), and trains the surrogate model $f_s$ used to craft the CPUPs (Section~\ref{sec:cpup}). Trainwreck attacks all classes in the dataset. For each attacked class $c_a$, the method first determines which images of $c_a$ are attacked and which stay clean. Trainwreck employs a simple strategy: it poisons the same ratio of images for each class, equal to the global poison rate $\pi$. For each class $c_a$, it uniformly randomly selects $\lfloor\pi n^{<c_a>}\rfloor$ images to be poisoned, leaving the remaining $\lceil(1-\pi)n^{<c_a>}\rceil$ images untouched. The index set of the poisoned images is denoted $\tilde{I}$. Next, the closest class $c_c$, $c_c \neq c_a$, with the minimum divergence from $c_a$ according to $\mathbf{D}$ is selected. Then, the CPUP for $c_a$ is computed. Note that the computation of CPUP uses \emph{all} the training data of class $c_a$, regardless of $\pi$. Finally, all images in $\tilde{I}$ get updated with the CPUP.

The output is the poisoned training dataset $\mathcal{D}_{tr}$ that for each attacked class $c_a$ shifts $\lfloor\pi n^{<c>}\rfloor$ images towards its closest class $c_c$, leaving the rest of the data and all the class labels intact. The method satisfies all three stealth criteria outlined in Section~\ref{sec:stealth}: the number of images in the dataset and per class remains the same, and the perturbation strength restriction is always met.
\section{Experimental setup}
\label{sec:exp_setup}

% Since DAAs have to the best of our knowledge not been systematically explored yet, there are no existing benchmarks. This section therefore proposes a challenging experimental setup to evaluate Trainwreck's claimed benefits. In particular, it aims to answer the following questions:

Bearing in mind the DAA requirements set in Section~\ref{sec:daa}, we answer the following questions to experimentally evaluate Trainwreck's performance:

\begin{itemize}
    \item (EQ1) Is Trainwreck a \emph{potent} DAA?
    \item (EQ2) Is Trainwreck a \emph{customizable} DAA?
    \item (EQ3) Is Trainwreck \emph{black-box} and \emph{transferable}?
    \item (EQ4) \emph{Ablation study}: Can Trainwreck be replaced by a subset of its components?
\end{itemize}

Note that stealth, the third and final DAA requirement, is not evaluated experimentally, as Trainwreck assures stealth by strictly adhering to the stealth objectives outlined in Section~\ref{sec:stealth}.

% EQ1 is straightforward, there should be evident considerable damage to the performance of models trained on Trainwreck-poisoned data. EQ2 is a sanity check: even in the absence of well-established baselines, Trainwreck must be pitted against sensible, intuitive techniques that an attacker would have used if tasked with a DAA. If Trainwreck yields the same performance as such techniques, then its value is low. EQ3 aims at the strength-stealth payoff: the attacker should be able to tune the strength to make the attack believable. This in turn makes the attack more difficult to discover and more damaging in the long run. Finally, answering EQ4 verifies whether Trainwreck truly is black-box, transferable, and suitable for usage across various datasets.

\subsection{Baselines}
\label{sec:baselides}

The experimental evaluation pits Trainwreck against seven baselines in total: three state-of-the-art data poisoning methods, three custom DAA baselines, and a random guess model (Random) as a reference frame. The three existing data poisoning approaches are learning to confuse (L2Confuse) by Feng \etal \cite{Feng19}, adversarial poisoning (AdvPoison) by Fowl \etal \cite{Fowl21}, and unlearnable examples (UnlearnEx) by Huang \etal \cite{Huang21}. All three recent data poisoning baselines use adversarial perturbation poison and operate under similar stealth requirements as Trainwreck, providing for a fair and challenging comparison

Since the existing data poisoning attacks cannot be compared to Trainwreck beyond EQ1, to provide more detailed results in DAA context and flesh out the answers to the remaining EQs, we introduce three custom baselines. Two of them are \emph{swapper} attacks that swap up to $\lfloor \frac{\pi n}{2} \rfloor$ training images between different classes, which ensures the poison rate $\pi$ is not exceeded. The first swapper attack is \emph{RandomSwap}, which simply performs uniformly randomly selected swaps. The second is \emph{JSDSwap}, a more ``informed'' swapper that uses the $\mathbf{D}$ matrix of Jensen-Shannon divergence (JSD) values (see Section~\ref{sec:dist_div}). JSDSwap goes down the list of classes ranked in ascending order by their minimal JSD from other classes, from each swapping $\lfloor \frac{\pi n^{<c>}}{2} \rfloor$ uniformly randomly selected images with images from the closest class according to $\mathbf{D}$.

% \begin{figure}
% \centering
% \includegraphics[width=0.8\columnwidth]{gfx/grid_stealth.pdf}
% \caption{Comparison of 2x4 image grids displaying CIFAR-10 data of class \emph{dog} poisoned by Trainwreck and a swapper attack (here: RandomSwap), both with $\pi=0.25$.}
% \label{fig:grid_stealth}
% \end{figure}

The final baseline, \emph{AdvReplace}, is a \emph{perturbation}-based attack that simply replaces a fraction of the training data proportional to $\pi$ with adversarial images resulting from running an \emph{untargeted} PGD attack on the surrogate model. Essentially, AdvReplace can be thought of as a generic, simple adversarial perturbation based poisoning attack. Comparing Trainwreck with JSDSwap and AdvReplace completes the ablation study, testing the importance of class divergence knowledge and adversarial perturbations without class knowledge, respectively.

\subsection{Datasets and parameters}
\label{sec:data_param}

The attack methods have been evaluated on CIFAR-10 and CIFAR-100 \cite{Krizhevsky09}, both established image classification benchmarks. The CIFAR datasets are simple: it is quite easy for modern models to learn to classify them them high accuracy. Precisely this makes them a challenging benchmark for DAAs, as it is hard to confuse the models with small perturbations or with swapping a few images.

The perturbation strength parameter $\epsilon$ is set to $\frac{8}{255}$ for all baselines. The poison rate $\pi$ does not need to change between datasets, but it does vary between swapper and perturbation attacks. Swapper attacks are visually conspicuous, explicitly breaking the perturbation stealth rule (see Equation~\ref{eq:stealth_pertstr}). To reflect this, we treat $\pi = 0.25$ as a generous upper bound for stealth for swapper attacks. The reason we err on the side of generosity is to present a challenge for Trainwreck by ensuring we do not disadvantage the swapper baselines. Therefore, we iterate over $\pi \in [0.05, 0.25]$ in swapper attack experiments. Perturbation attacks iterate over$\pi \in [0.25, 1]$: $\pi=1$ is permissible due to the full adherence to the stealth requirements, while $\pi < 0.25$ would result in minimal to no effect on the target model due to the perturbations being small. For the other Trainwreck and custom baselines parameters, we use the empirical values as described in Section~\ref{alg:cpup}: $n_{iter}^{cpup}=1$, $n_{iter}^{pgd}=10$.

\subsection{Evaluation protocol}
\label{sec:eval_protocol}

We have run Trainwreck and custom baseline attacks on three target model architectures. For each evaluated method, we report the attacked model's \emph{test top-1 accuracy} on the original test dataset $\mathcal{D}_{ts}$ (note that none of the attack methods tamper with $\mathcal{D}_{ts}$). To eliminate early-epoch randomness, the highest accuracy value in the final 10 epochs is reported. The damage to test accuracy is a measure of the evaluated DAA's success, in line with Section~\ref{sec:daa}.

The three target model architectures evaluated in our experiments are EfficientNetV2 (EN-V2) \cite{Tan21}, ResNeXt-101 (RN-101) \cite{Xie17}, and the ViT-L-16 vision transformer (FT-ViT) \cite{Kolesnikov21}. As the surrogate model, Trainwreck and AdvReplace use ResNet-50 \cite{He16}. All EN-V2 and RN-101 models were trained from scratch, all transformer models had been initialized to pre-trained ImageNet weights and finetuned: only the new fully connected output layer is trained, the rest of the parameters are frozen. All models have been trained for 30 epochs: not enough for full performance, but sufficient for the attack effects to converge.

This choice of models allows evaluating a number of scenarios. Firstly, the surrogate model's architecture is less advanced than the targets': in a real setting, the attacker likely has a weaker model than the target due to lower domain knowledge. This tests realistic \emph{transferability}. To test the \emph{black-box} claim, one of the target architectures, RN-101, is more similar to the surrogate model than the rest. If RN-101 results are noticeably better, then it is doubtful the attack is black-box. Finally, the difficulty of attacking FT-ViT should not be understated: it entails manipulating a large model with $\sim$300M parameters, pretrained on a rich image dataset, while poisoning a single output layer.
\section{Experimental results}
\label{sec:exp_results}

\begin{figure}
\centering
\includegraphics[width=\textwidth]{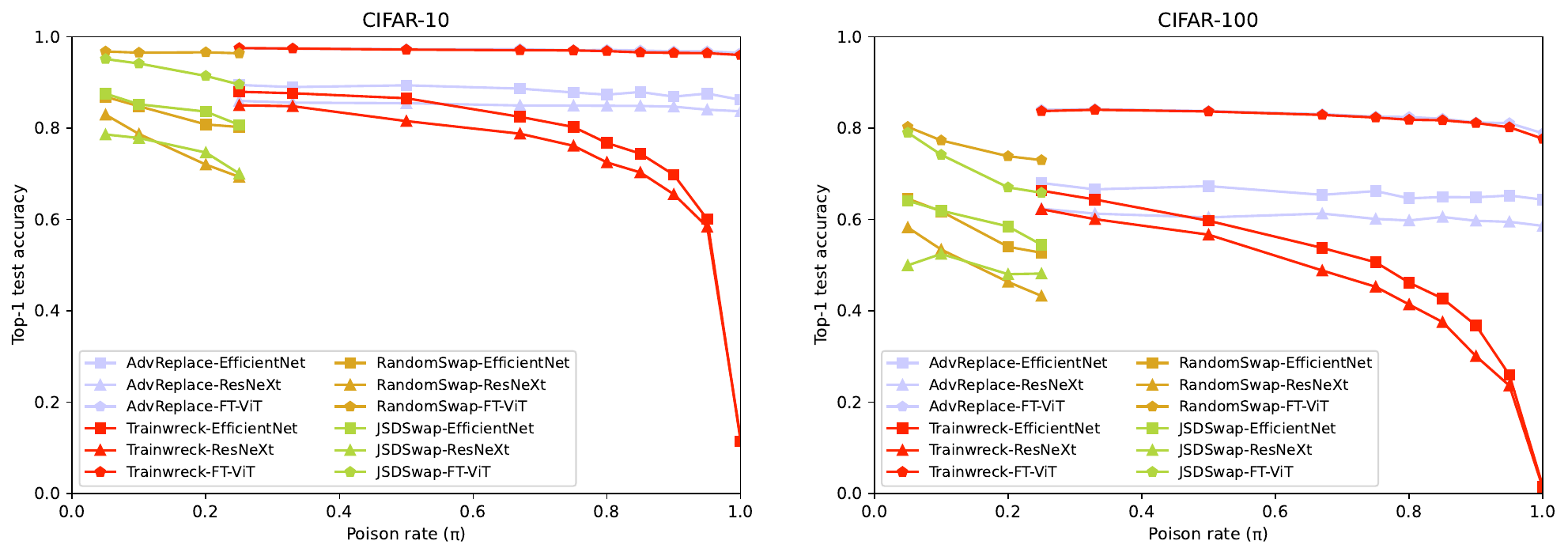}
\caption{Test top-1 accuracy results for varying poison rate $\pi$.}
\label{fig:poison_plot}
\end{figure}

Table~\ref{tab:comparison} reports the strongest poisoning results achieved by Trainwreck and the existing state-of-the-art poisoning attacks. Table~\ref{tab:results} reports the test top-1 accuracy for the strongest versions of Trainwreck and the custom baselines. For the sake of completeness, the test top-1 accuracy of the surrogate ResNet-50 model is 0.8557 on CIFAR-10 and 0.5923 on CIFAR-100, both lower than the respective target models. Figure~\ref{fig:poison_plot} plots test top-1 accuracy for varying $\pi$.

\begin{table}
\begin{center}
\begin{tabular}{|lll|rr|}
    \hline
    
    \textbf{Attack} & \textbf{Surrogate arch.} & \textbf{Target arch.} & \textbf{CIFAR-10} & \textbf{CIFAR-100}\\
    \hline
    Random & N/A & N/A & 0.1 & 0.01 \\
    L2Confuse \cite{Feng19} & U-Net & ConvNet & 0.2877 & N/A \\
    AdvPoison \cite{Fowl21} & ResNet-18 & ResNet-18 & 0.0625 & N/A \\
    UnlearnEx \cite{Huang21} & ResNet-18 &  ResNet-50 & 0.1345 & 0.0380 \\
    Trainwreck ($\pi = 1)$ & ResNet-50 &  ResNeXt-101 & 0.1211 & 0.0135 \\
    \hline
\end{tabular}
\end{center}
\caption{Test accuracy of models damaged by train data poisoning. For each attack, the lowest achieved accuracy and the corresponding architectures are reported.}
\label{tab:comparison}
\end{table}

\begin{table}
\begin{center}
\begin{tabular}{|l|rrr|rrr|}
    \hline
    & \multicolumn{3}{c|}{\textbf{CIFAR-10}} & \multicolumn{3}{c|}{\textbf{CIFAR-100}} \\
    \textbf{Method} & \textbf{EN-V2} & \textbf{RN-101} & \textbf{FT-ViT} & \textbf{EN-V2} & \textbf{RN-101} & \textbf{FT-ViT}\\ \hline
    Clean model & 0.8962 & 0.8808 & 0.9795 & 0.6876 & 0.6309 & 0.8563 \\
    RandomSwap ($\pi=0.25$) & 0.8115 & 0.6933 & 0.9638 & 0.5272 & 0.4327 & 0.7366 \\
    JSDSwap ($\pi=0.25$) & 0.8071 & 0.7005 & \textbf{0.8965} & 0.5447 & 0.4816 & \textbf{0.6585} \\
    AdvReplace ($\pi=1$) & 0.8620 & 0.8369 & 0.9648 & 0.6435 & 0.5860 & 0.7930 \\
    Trainwreck ($\pi=1$) & \textbf{0.1146} & \textbf{0.1211} & 0.9600 & \textbf{0.0157} & \textbf{0.0135} & 0.7788 \\
    \hline
\end{tabular}
\end{center}
\caption{Test top-1 accuracy results of Trainwreck and the custom baselines.}
\label{tab:results}
\end{table}

The answer to EQ1 is that Trainwreck indeed is a \emph{potent} attack. Within 30 epochs, the target models trained from scratch on Trainwreck-attacked data permanently degrade to a performance barely above random choice. As Table~\ref{tab:comparison} shows, this potency is comparable or even better than the data poisoning state of the art. On the difficult FT-ViT model, the accuracy drop caused by Trainwreck is several percent (1.95\% on CIFAR-10 and 7.75\% on CIFAR-100). This accuracy drop is smaller, but a noticeable adversarial effect is still present, even though the attack could only influence the final output layer parameters with stealthy perturbations.

Figure~\ref{fig:poison_plot} convincingly answers EQ2: Trainwreck is a \emph{customizable} DAA. For both datasets, the relationship between the test accuracy and $\pi$ can be decomposed roughly into three regions. For $\pi < 0.5$, the attacks have a limited effect, $\pi \in [0.5, 0.8]$ results in a moderate drop in performance, and for $\pi > 0.8$, the performance starts degrading rapidly. Since the attacker can choose $\pi$ freely and  thus influence the attack strength, Trainwreck is indeed customizable. Note that none of the data poisoning baselines are customizable: L2Confuse \cite{Feng19} and UnlearnEx \cite{Huang21} require full poisoning ($\pi = 1$), the AdvPoison paper reports their partial poisoning merely ``does not significantly improve results over training on only the clean data and often degrades results below what one would achieve using only the clean data'' \cite{Fowl21}. Customizability is therefore, to the best of our knowledge, unique to Trainwreck.

To answer EQ3: the black-box claim is verified by the absence of a performance spike on ResNeXt-101. The surrogate model with a similar architecture does not have a noticeable advantage, \ie, the attacker does not have to match the target model's architecture. Transferability is verified by the single poisoning of each dataset resulting in a degraded performance across target models. Some variance in attack strength is present due to FT-ViTs being difficult to attack by nature in this setting. Overall, Trainwreck is black-box and transferable.

To answer EQ4, overall, Trainwreck's components, JSDSwap and AdvReplace, are not an adequate substitute for Trainwreck, although they can be a better choice in certain scenarios. In particular, the FT-ViT model seems to respond more strongly to swapper attacks than to Trainwreck. Looking at the results closer, a likely explanation is that the stealthy perturbations of Trainwreck simply do not have the ``gradient power'' to sway the mostly-frozen FT-ViT model during finetuning. The deciding factor therefore are JSDSwap's more drastic data edits. On all other models, Trainwreck is clearly the superior choice.

\section{Discussion \& defense}
\label{sec:discussion}

Trainwreck proving its mettle as a pioneer DAA may lead to serious consequences for organizations using AI. This section discusses the potential negative impacts and methods to reliably defend against Trainwreck. Trainwreck makes it difficult to pinpoint the poisoned dataset as the source of the low performance. Trainwreck's stealth requirements ensure that even the strongest version of the attack is difficult to see if focusing on one image at a time, let alone glancing at an image grid in a file manager. If the attack strength is customized to result in plausible performance degradation, the model stakeholders may attribute the low performance to other reasons, \eg, wrong model architecture, training procedure, or low quality of data. Trainwreck can therefore leave lingering damage.

The first defense imperative is \emph{data redundancy}. Trainwreck or another train-time DAA is especially devastating if the single existing copy of the data is poisoned. The defenders should therefore ensure a guaranteed-clean \emph{canonical} copy of the data. This canonical copy should be stored in multiple locations with separate access and sound access policies in place. With a canonical copy of the data, \emph{data hashing} is a straightforward, reliable defense method---provided ed a sufficiently strong hash that resists brute-force attacks is used, such as SHA-256 or SHA-512. For weaker hashes such as MD5, it may be easy to alter poisoned data with file content prefixes or suffixes to match the hash of the original data \cite{Stevens12}. If the defenders have sufficiently strong canonical data hashes, all they need to do to reliably detect a train-time DAA is to compare the hashes of the data in the training dataset with the canonical ones.
\section{Conclusion}
\label{sec:conc}

This paper opens up the topic of damaging adversarial attacks (DAAs) that explicitly damage computer vision models, formalizing the DAA definition, threat model, DAA objectives, and three requirements for a successful DAA: potency, stealth, and customizability. The Trainwreck attack proposed in this paper meets all three criteria, making it a pioneer train-time DAA on image classifiers. Trainwreck is also a black-box, transferable attack.

With the importance of computer vision models in applied practice increasing, DAAs are emerging as impactful economical sabotage tools. It is high time they receive proper attention. Beyond contributing the open-source code for the Trainwreck attack, this paper proposes reliable defense methods against train-time DAAs. We hope that this contribution fosters future research on DAAs.

\bibliographystyle{splncs04}
\bibliography{main}

\begin{thebibliography}{10}
\providecommand{\url}[1]{\texttt{#1}}
\providecommand{\urlprefix}{URL }
\providecommand{\doi}[1]{https://doi.org/#1}

\bibitem{Mitre23}
ATT\&CK, M.: Tactics: Impact. \url{https://attack.mitre.org/tactics/TA0040/} (2023), [Online; accessed 9 Nov 2023]

\bibitem{Cai22}
Cai, Z., Rane, S., Brito, A.E., Song, C., Krishnamurthy, S.V., Roy-Chowdhury, A.K., Asif, M.S.: Zero-query transfer attacks on context-aware object detectors. In: CVPR. pp. 15004--15014 (2022). \doi{10.1109/CVPR52688.2022.01460}

\bibitem{Chen23}
Chen, S., Chen, H., Haque, M., Liu, C., Yang, W.: The dark side of dynamic routing neural networks: Towards efficiency backdoor injection. In: CVPR. pp. 24585--24594 (2023). \doi{10.1109/CVPR52729.2023.02355}

\bibitem{Chen22}
Chen, S., Song, Z., Haque, M., Liu, C., Yang, W.: {NICGS}low{D}own: Evaluating the efficiency robustness of neural image caption generation models. In: CVPR. pp. 15344--15353 (2022). \doi{10.1109/CVPR52688.2022.01493}

\bibitem{ChenZ22}
Chen, Z., Li, B., Xu, J., Wu, S., Ding, S., Zhang, W.: Towards practical certifiable patch defense with vision transformer. In: CVPR. pp. 15127--15137 (2022). \doi{10.1109/CVPR52688.2022.01472}

\bibitem{Feng19}
Feng, J., Cai, Q.Z., Zhou, Z.H.: Learning to confuse: generating training time adversarial data with auto-encoder. NeurIPS  \textbf{32} (2019)

\bibitem{Feng23}
Feng, W., Xu, N., Zhang, T., Zhang, Y.: Dynamic generative targeted attacks with pattern injection. In: CVPR. pp. 16404--16414 (2023). \doi{10.1109/CVPR52729.2023.01574}

\bibitem{Fowl21}
Fowl, L., Goldblum, M., Chiang, P.y., Geiping, J., Czaja, W., Goldstein, T.: Adversarial examples make strong poisons. In: NeurIPS. vol.~34, pp. 30339--30351 (2021)

\bibitem{Goodfellow15}
Goodfellow, I., Shlens, J., Szegedy, C.: Explaining and harnessing adversarial examples. In: ICLR (2015), \url{http://arxiv.org/abs/1412.6572}

\bibitem{Gowal21}
Gowal, S., Rebuffi, S.A., Wiles, O., Stimberg, F., Calian, D.A., Mann, T.A.: Improving robustness using generated data. In: Ranzato, M., Beygelzimer, A., Dauphin, Y., Liang, P., Vaughan, J.W. (eds.) NeurIPS. vol.~34, pp. 4218--4233. Curran Associates, Inc. (2021), \url{https://proceedings.neurips.cc/paper_files/paper/2021/file/21ca6d0cf2f25c4dbb35d8dc0b679c3f-Paper.pdf}

\bibitem{Gu17}
Gu, T., Dolan{-}Gavitt, B., Garg, S.: Badnets: Identifying vulnerabilities in the machine learning model supply chain. CoRR  \textbf{abs/1708.06733} (2017), \url{http://arxiv.org/abs/1708.06733}

\bibitem{He16}
He, K., Zhang, X., Ren, S., Sun, J.: Deep residual learning for image recognition. In: 2016 IEEE Conference on Computer Vision and Pattern Recognition (CVPR). pp. 770--778 (2016). \doi{10.1109/CVPR.2016.90}

\bibitem{Hu21}
Hu, Y.C.T., Kung, B.H., Tan, D.S., Chen, J.C., Hua, K.L., Cheng, W.H.: Naturalistic physical adversarial patch for object detectors. In: ICCV. pp. 7848--7857 (October 2021)

\bibitem{Hu23}
Hu, Z., Chu, W., Zhu, X., Zhang, H., Zhang, B., Hu, X.: Physically realizable natural-looking clothing textures evade person detectors via 3d modeling. In: CVPR. pp. 16975--16984 (2023). \doi{10.1109/CVPR52729.2023.01628}

\bibitem{Huang21}
Huang, H., Ma, X., Erfani, S.M., Bailey, J., Wang, Y.: Unlearnable examples: Making personal data unexploitable. In: ICLR (2021)

\bibitem{Huang23}
Huang, H., Chen, Z., Chen, H., Wang, Y., Zhang, K.: T-sea: Transfer-based self-ensemble attack on object detection. In: CVPR. pp. 20514--20523 (2023). \doi{10.1109/CVPR52729.2023.01965}

\bibitem{Huang22}
Huang, Q., Dong, X., Chen, D., Zhou, H., Zhang, W., Yu, N.: Shape-invariant 3d adversarial point clouds. In: CVPR. pp. 15314--15323 (2022). \doi{10.1109/CVPR52688.2022.01490}

\bibitem{HuangS23}
Huang, S., Lu, Z., Deb, K., Boddeti, V.N.: Revisiting residual networks for adversarial robustness. In: CVPR. pp. 8202--8211 (2023). \doi{10.1109/CVPR52729.2023.00793}

\bibitem{Jiang23}
Jiang, W., Li, H., Xu, G., Zhang, T.: Color backdoor: A robust poisoning attack in color space. In: CVPR. pp. 8133--8142 (2023). \doi{10.1109/CVPR52729.2023.00786}

\bibitem{koh17}
Koh, P.W., Liang, P.: Understanding black-box predictions via influence functions. In: Int. Conf. Machine Learning. pp. 1885--1894. PMLR (2017)

\bibitem{Kolesnikov21}
Kolesnikov, A., Dosovitskiy, A., Weissenborn, D., Heigold, G., Uszkoreit, J., Beyer, L., Minderer, M., Dehghani, M., Houlsby, N., Gelly, S., Unterthiner, T., Zhai, X.: An image is worth 16x16 words: Transformers for image recognition at scale. In: ICLR (2021)

\bibitem{Krizhevsky09}
Krizhevsky, A.: Learning multiple layers of features from tiny images. Tech. rep., University of Toronto (2009)

\bibitem{Kurakin17}
Kurakin, A., Goodfellow, I.J., Bengio, S.: Adversarial machine learning at scale. In: ICLR (2017), \url{https://arxiv.org/abs/1611.01236}

\bibitem{kushner2013stuxnet}
Kushner, D.: The real story of {S}tuxnet. IEEE Spectrum  \textbf{50}(3),  48--53 (2013)

\bibitem{Li23}
Li, Y., Li, Y., Dai, X., Guo, S., Xiao, B.: Physical-world optical adversarial attacks on 3d face recognition. In: CVPR. pp. 24699--24708 (2023). \doi{10.1109/CVPR52729.2023.02366}

\bibitem{Liu20}
Liu, Y., Ma, X., Bailey, J., Lu, F.: Reflection backdoor: A natural backdoor attack on deep neural networks. In: Vedaldi, A., Bischof, H., Brox, T., Frahm, J.M. (eds.) ECCV. pp. 182--199. Springer International Publishing, Cham (2020)

\bibitem{Madry18}
Madry, A., Makelov, A., Schmidt, L., Tsipras, D., Vladu, A.: Towards deep learning models resistant to adversarial attacks. In: ICLR (2018), \url{https://arxiv.org/abs/1706.06083}

\bibitem{MoosaviDezfooli17}
Moosavi-Dezfooli, S.M., Fawzi, A., Fawzi, O., Frossard, P.: Universal adversarial perturbations. In: CVPR. pp. 86--94 (2017). \doi{10.1109/CVPR.2017.17}

\bibitem{Rebuffi21}
Rebuffi, S.A., Gowal, S., Calian, D.A., Stimberg, F., Wiles, O., Mann, T.A.: Data augmentation can improve robustness. In: Ranzato, M., Beygelzimer, A., Dauphin, Y., Liang, P., Vaughan, J.W. (eds.) NeurIPS. vol.~34, pp. 29935--29948. Curran Associates, Inc. (2021), \url{https://proceedings.neurips.cc/paper_files/paper/2021/file/fb4c48608ce8825b558ccf07169a3421-Paper.pdf}

\bibitem{Rony23}
Rony, J., Pesquet, J.C., Ayed, I.B.: Proximal splitting adversarial attack for semantic segmentation. In: CVPR. pp. 20524--20533 (2023). \doi{10.1109/CVPR52729.2023.01966}

\bibitem{Saha22}
Saha, A., Tejankar, A., Koohpayegani, S.A., Pirsiavash, H.: Backdoor attacks on self-supervised learning. In: CVPR. pp. 13337--13346 (June 2022)

\bibitem{Salman22}
Salman, H., Jain, S., Wong, E., Mądry, A.: Certified patch robustness via smoothed vision transformers. In: CVPR. pp. 15116--15126 (2022). \doi{10.1109/CVPR52688.2022.01471}

\bibitem{Stevens12}
Stevens, M., Lenstra, A., Weger, B.: Chosen-prefix collisions for md5 and applications. International Journal of Applied Cryptography  \textbf{2} (07 2012). \doi{10.1504/IJACT.2012.048084}

\bibitem{Szegedy14}
Szegedy, C., Zaremba, W., Sutskever, I., Bruna, J., Erhan, D., Goodfellow, I., Fergus, R.: Intriguing properties of neural networks. In: ICLR (2014), \url{http://arxiv.org/abs/1312.6199}

\bibitem{Tan21}
Tan, M., Le, Q.V.: {EfficientNetV2}: Smaller models and faster training. In: Int. Conf. Machine Learning (2021), \url{https://arxiv.org/pdf/2104.00298.pdf}

\bibitem{Walmer22}
Walmer, M., Sikka, K., Sur, I., Shrivastava, A., Jha, S.: Dual-key multimodal backdoors for visual question answering. In: 2022 IEEE/CVF Conference on Computer Vision and Pattern Recognition (CVPR). pp. 15354--15364 (2022). \doi{10.1109/CVPR52688.2022.01494}

\bibitem{Williams23}
Williams, P.N., Li, K.: Black-box sparse adversarial attack via multi-objective optimisation cvpr proceedings. In: CVPR. pp. 12291--12301 (2023). \doi{10.1109/CVPR52729.2023.01183}

\bibitem{Xie17}
Xie, S., Girshick, R., Dollár, P., Tu, Z., He, K.: Aggregated residual transformations for deep neural networks. In: CVPR. pp. 5987--5995 (2017). \doi{10.1109/CVPR.2017.634}

\bibitem{Yang23}
Yang, X., Liu, C., Xu, L., Wang, Y., Dong, Y., Chen, N., Su, H., Zhu, J.: Towards effective adversarial textured 3d meshes on physical face recognition. In: CVPR. pp. 4119--4128 (2023). \doi{10.1109/CVPR52729.2023.00401}

\bibitem{Zhang21}
Zhang, B., Cai, T., Lu, Z., He, D., Wang, L.: Towards certifying l-infinity robustness using neural networks with l-inf-dist neurons. In: Meila, M., Zhang, T. (eds.) Proceedings of the 38th International Conference on Machine Learning. Proceedings of Machine Learning Research, vol.~139, pp. 12368--12379. PMLR (18--24 Jul 2021), \url{https://proceedings.mlr.press/v139/zhang21b.html}

\bibitem{Zhong22}
Zhong, Y., Liu, X., Zhai, D., Jiang, J., Ji, X.: Shadows can be dangerous: Stealthy and effective physical-world adversarial attack by natural phenomenon. In: 2022 IEEE/CVF Conference on Computer Vision and Pattern Recognition (CVPR). pp. 15324--15333 (2022). \doi{10.1109/CVPR52688.2022.01491}

\end{thebibliography}
\end{document}